\documentclass[conference]{IEEEtran}

\usepackage{datetime}
\usepackage{url}
\usepackage{hyperref}

\usepackage{graphicx}
\DeclareGraphicsExtensions{.pdf,.jpeg,.png}

\begin{document}

\title{User-centric Composable Services: A New Generation of Personal Data Analytics}

\author{\IEEEauthorblockN{Jianxin Zhao,
Richard Mortier, Jon Crowcroft and
Liang Wang}
\IEEEauthorblockA{Computer Laboratory,
University of Cambridge\\
Email: \{first.last\}@cl.cam.ac.uk}}

\maketitle

\begin{abstract}
Machine Learning (ML) techniques, such as Neural Network, are widely used in today's applications. However, there is still a big gap between the current ML systems and users' requirements. ML systems focus on improving the performance of models in training, while individual users cares more about response time and expressiveness of the tool. Many existing research and product begin to move computation towards edge devices. Based on the numerical computing system Owl, we propose to build the Zoo system to support construction, compose, and deployment of ML models on edge and local devices.
\end{abstract}

\section{Introduction}

Nowadays Machine Learning techniques such as Deep Neural Network (DNN) are used in numerous services.
However, there is a big gap between the current ML systems and users' requirements.
On one hand, most existing machine learning frameworks, such as TensorFlow and Caffe, focus mainly on the ``training'' phase.
They aim at accelerating the training speed, enhancing performance on GPU, or improving prediction accuracy.
On the other hand however, the users, either individuals who want to use the ML-based services or researchers who do not fully commit to the ML field, care less about those benchmarks, but rather about issues such as expressiveness of the tool for constructing a neural network, fast development of new algorithms or neurons on existing systems, access to ML models on local devices, service response time, etc.

Both academia and industry begin to mitigate this gap. One crucial aspect of existing solutions is to move ML computation towards local devices.
Most current end-side services, such as personal intelligent assistants and smart home service, either only support simple ML models or require users to upload raw data (speech, image, etc.) to complex data analytics services host on the cloud.
The latter practice is known to associate with issues such as communication cost, latency, and personal data privacy.
Neurosurgeon \cite{kang2017neurosurgeon} is a system to partition a Deep Neural Network into two parts, half on edge devices and the other half on cloud, to reduce computing latency and energy consumption.
Similarly, \cite{teerapittayanon2017distributed} also propose to partition a neuron network across mobile devices, edges, and cloud, so as to give results with lower latency.
\cite{McMahanMRA16} proposes an algorithm in the collaborative training of a model by multiple mobile devices.
\cite{RodriguezWZMH17} explore the method of training personalised model on local devices from an initial shared model, to provide for model training and inference in a system where computation is moved to the data.
\cite{wang2016kvasir} focus on a particular kind of data analytics: semantic analysis and recommendation of related articles from large text corpora. It pushes this content provision service into web browser at the users' end.
Some systems also begin to focus on mobile platforms, such as Facebook's \texttt{Caffe2go}, but they still place emphasis on shifting what existing computing platform can do from data centre to mobile devices, and have not provided systematic solutions to address the aforementioned issues.


In this poster, we present overall design of Owl system, its advantages over other learning platforms, and propose a ``Zoo'' module built on Owl to mitigate this gap. It aims to provide the whole lifetime of a ML service, from model construction, compose, and deployment on edge and local devices.

\section{System Architecture}

\begin{figure}[!t]
    \centering
    \includegraphics[width=\columnwidth]{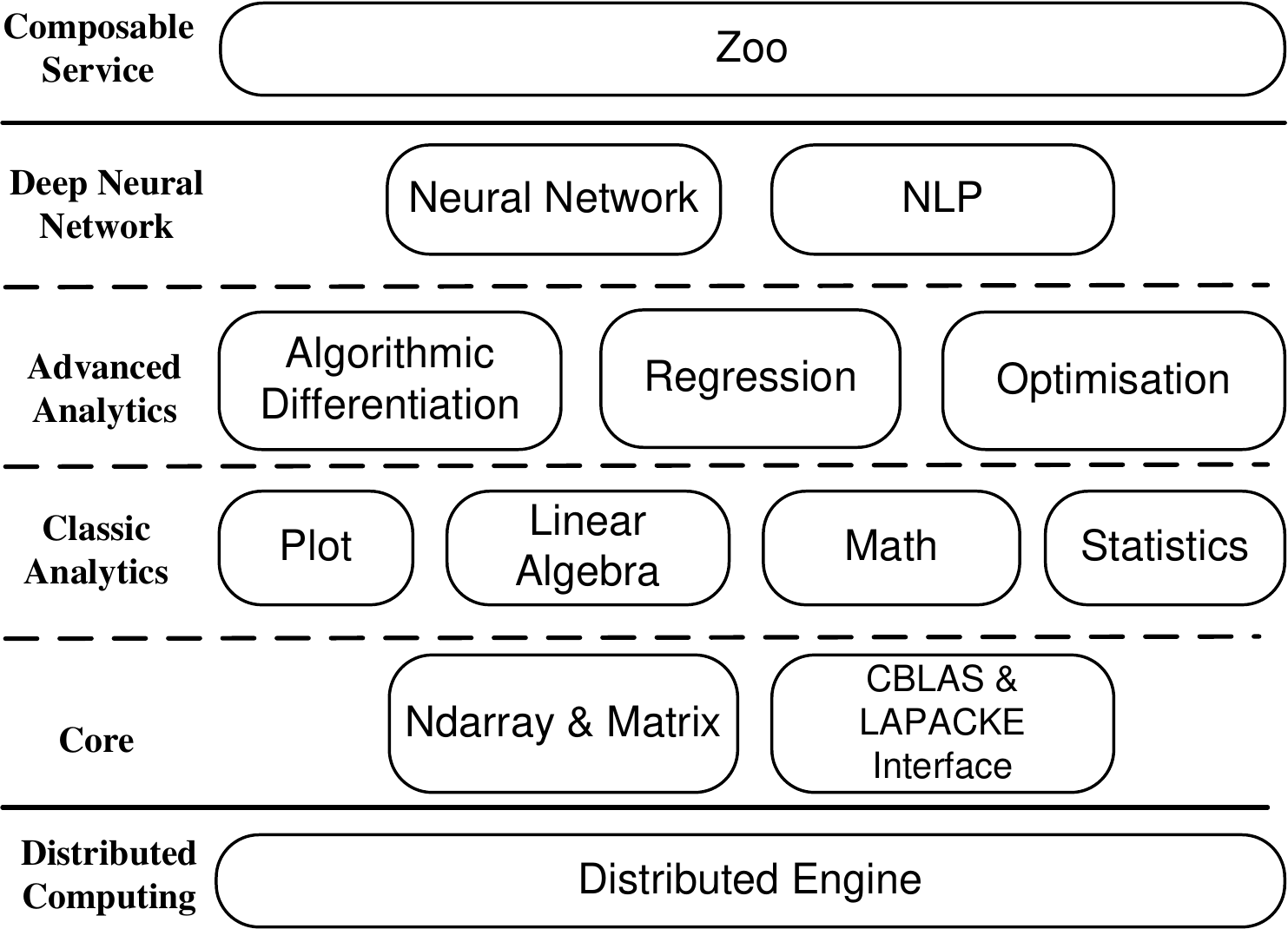}
    \caption{Owl Architecture}
    \label{fig:owl}
  \vspace{-1.4em}
\end{figure}

Owl \cite{liang2017owl} is an open-source numerical computing system in OCaml language.
Owl provides a full stack support for numerical methods, scientific computing, and advanced data analytics on OCaml.
Figure~\ref{fig:owl} shows the architecture of Owl system.
Built on the core data structure of matrix and n-dimensional array, Owl supports a comprehensive set of classic analytics such as math functions, statistics, linear algebra, as well as advanced analytics techniques, namely optimisation, algorithmic differentiation, and regression.
On top of them, Owl provides Neural Network (NN) and Natural Language Processing (NLP) modules.

Owl system can be extended towards two directions.
First, it can use the parallel and distributed engine at lower level to support distributed numerical computing and data analytics. It supports different protocols and multiple barrier control techniques\cite{liang2017prop}.
Second, based on the ML modules, Owl can connect to Zoo, a module that support \textit{Composable Services}.
Its basic idea is users should not have to construct new ML services every time new application requirements arise.
In fact, many services can be composed from basic ML services: image recognition, speech-to-text, recommendation, etc.
The Zoo module aims at providing user-centric, ML-based services, enabling service pulling, sharing, compatibility checking, and composing on local devices.

\section{Initial Performance Evaluation}

Next, I present preliminary experiment results on Owl, especially its Neural Network module, since it is the most important module that the zoo module is based upon.

The most exciting feature of Owl is its expressiveness.
We have constructed InceptionV3 model\cite{chris2015rethinking}, one of the most complex network architecture in existing image recognition models, with only 150 LoC, while constructing the same model requires 400 LoC using TensorFlow code.
Besides enabling shorter and more compact code, another of its advantages compared with existing popular learning platforms is its flexibility to add new features.
As an example, we insert instrumentation code into Owl to collect the computing latency of each node in a neural network when doing inference. Adding this feature only takes 50 LoC.
Our initial experiment shows an acceptable performance tradeoff, which is only about 2 times slower than state-of-the-art TensorFlow and Caffe2.

Besides enabling shorter and more compact code, Owl also supports fast development of new features.
As an example, I've inserted instrumenting code into Owl's source code to collect the computing latency of each node in a neural network when doing inference. And all it takes is only50 LoC.
Using this newly added feature, I get the computing latency of each node in the Inception neural network.
This new feature proves to provides deep insights into how each node works and guidance on possible optimisations.

With great expressiveness comes some performance tradeoff. Currently Owl is not as fast as industry-level mature products such as TensorFlow and Caffe2. The question is: how much slower?
Since the Zoo relies heavily on inference using Owl's NN module, we want to compare the inference time on Owl and the other two platforms.
We choose three representative DNNs that vary greatly in architecture complexity and parameter sizes:
1) one small neural network (LeNet-5~\cite{lecun1998gradient}) that only consists of 8 nodes and contains about 240KB parameters, for the MNIST handwriting recognition task;
2) an VGG16~\cite{karenvgg14} model that has a simple architecture with 38 nodes but a large amount of parameters (500MB) for real-world image recognition task;
3) an InceptionV3 model also for image recognition, with less parameters (100MB), but a far more complex architecture (313 nodes).
We compare the time it takes for each model to finish its inference task on different platforms: Owl, TensorFlow, and Caffe2.
We use Zoo to deploy these models on Owl.

\begin{figure}[!t]
    \centering
    \includegraphics[width=\columnwidth]{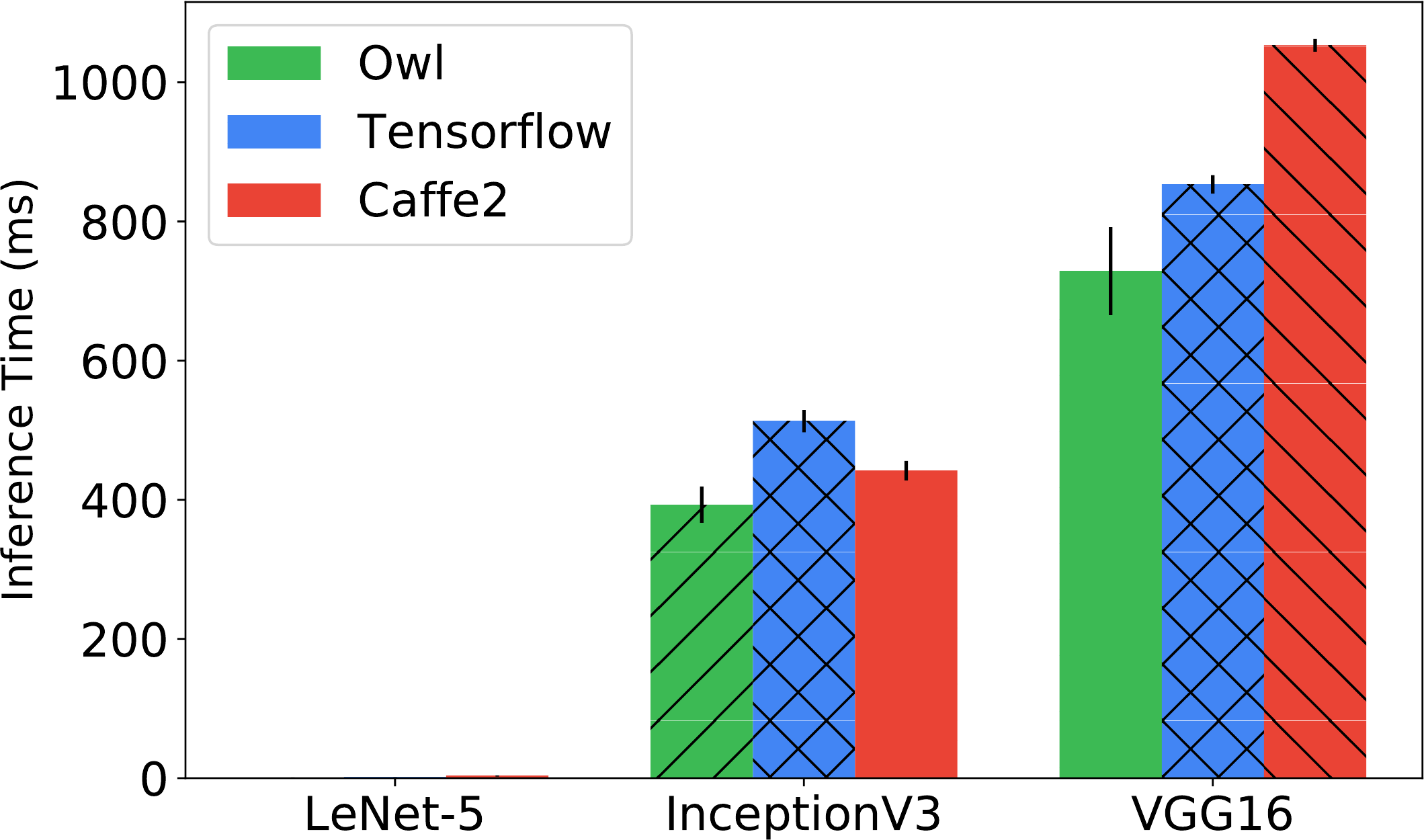}
    \caption{Inference time comparison on different platforms.}
    \label{fig:time}
    \vspace{-1em}
\end{figure}

The results are shown in Figure~\ref{fig:time}.
We can see that, regardless of great diversities in these models' architectures and sizes, Owl takes less time to do inference than Tensorflow and Caffe2. It means that Owl can achieve both expressiveness and good preformance.
The superior performance of Owl on large models is attributed to its efficient math operations.

\section{Summary}

In summary, based on Owl system, we are proposing to build the Zoo system. It aims to mitigate the current gap between current ML computing systems and users' requirements.
Initial experiment on Owl proves its outstanding expressiveness and acceptable performance tradeoff.
We believe this area of research is only just beginning to gain momentum.

\textbf{Acknowledgments}
This work is funded in part by the EPSRC Databox project (EP/N028260/1), NaaS (EP/K031724/2) and Contrive (EP/N028422/1).

\bibliographystyle{unsrt}
\bibliography{sosp}

\end{document}